\documentclass[sigconf,language=english]{acmart} 
\usepackage{graphicx} 
\usepackage{booktabs} 
\usepackage{multirow}
\usepackage{hyperref}
\usepackage{verbatim}

\usepackage{CJKutf8}

\AtBeginDocument{%
  }

\setcopyright{acmlicensed}
\copyrightyear{2025}
\acmYear{2025}
\acmDOI{XXXXXXX.XXXXXXX}
\acmConference[Agentic \& GenAI Evaluation Workshop KDD'25]{KDD workshop on Evaluation and Trustworthiness of Agentic and Generative AI Models}{August 4,
  2025}{Toronto, ON, Canada}
\acmISBN{978-1-4503-XXXX-X/2025/08}

\begin{document}

\title{Uncovering the Fragility of Trustworthy LLMs through Chinese Textual Ambiguity}

\author{Xinwei Wu}
\authornote{These authors contributed equally to this research.}
\author{Hongyu Liu}
\authornotemark[1]
\author{Haojie Li}
\authornotemark[1]
\affiliation{%
  \institution{Chalmers University of Technology}
  \city{Gothenburg}
  \country{Sweden}
}

\author{Xinyu Ji}
\affiliation{%
  \institution{Boeing}
  \city{Gothenburg}
  \country{Sweden}
}

\author{Ruohan Li}
\affiliation{%
  \institution{Chalmers University of Technology}
  \city{Gothenburg}
  \country{Sweden}
}

\author{Yule Chen}
\affiliation{%
  \institution{Chalmers University of Technology}
  \city{Gothenburg}
  \country{Sweden}
}

\author{Yigeng Zhang}
\authornote{Corresponding author. Email: yzhang168@uh.edu}
\affiliation{%
  \institution{\vspace{\baselineskip}}
  \city{ }
  \country{USA}}

\renewcommand{\shortauthors}{Wu et al.}

\begin{abstract}
In this work, we study a critical research problem regarding the trustworthiness of large language models (LLMs): how LLMs behave when encountering ambiguous narrative text, with a particular focus on Chinese textual ambiguity. We created a benchmark dataset by collecting and generating ambiguous sentences with context and their corresponding disambiguated pairs, representing multiple possible interpretations. These annotated examples are systematically categorized into 3 main categories and 9 subcategories.
Through experiments, we discovered significant fragility in LLMs when handling ambiguity, revealing behavior that differs substantially from humans. Specifically, LLMs cannot reliably distinguish ambiguous text from unambiguous text, show overconfidence in interpreting ambiguous text as having a single meaning rather than multiple meanings, and exhibit overthinking when attempting to understand the various possible meanings.
Our findings highlight a fundamental limitation in current LLMs that has significant implications for their deployment in real-world applications where linguistic ambiguity is common, calling for improved approaches to handle uncertainty in language understanding. 
The dataset and code are publicly available at \href{https://github.com/ictup/LLM-Chinese-Textual-Disambiguation}{this GitHub repository}\footnote{\url{https://github.com/ictup/LLM-Chinese-Textual-Disambiguation}}.

\end{abstract}




\maketitle

\section{Introduction}
Large Language Models (LLMs) have demonstrated strong language understanding capabilities and are widely deployed across a range of real-world applications \cite{li2023large, thirunavukarasu2023large, raiaan2024review}. They are used to process complex instructions in multi-turn dialogues and are integrated into various systems as agents or components of AI workflows. However, LLMs still exhibit inherent limitations in trustworthiness, such as hallucinations \cite{azamfirei2023large, huang2025survey}, misunderstanding \cite{liu2023we}, and misalignment \cite{dung2023current} that are particularly critical in safety-sensitive scenarios. Researchers have also invested significant effort in improving alignment, developing guardrails \cite{ayyamperumal2024current}, and enhancing uncertainty understanding \cite{he2023survey, shorinwa2025survey} to enable more reliable use of LLMs.

In practical use cases, people typically interact with LLMs via chat interfaces using written text in a conversational or spoken style, where ambiguity frequently arises. For example, in an LLM-based e-commerce shopper agent, the instruction \emph{return the phone and computer accessories I purchased last month} is ambiguous: it could mean returning the phone and the computer's accessories, or the accessories for both the phone and the computer. In such cases, the agent should be able to use appropriate means to resolve the ambiguity instead of proceeding with one possible interpretation, which may lead to unintended outcomes.

In this work, we focus specifically on examining how LLMs behave when faced with linguistic ambiguity, a core aspect of human language understanding. We present a new benchmark for ambiguity detection and interpretation in Chinese text. The dataset was annotated by native Chinese speakers and includes ambiguity type classification. It contains 900 ambiguous sentences sourced from real-world contexts spanning a variety of everyday scenarios. Each ambiguous sentence is annotated with all plausible interpretations and a corresponding set of disambiguated sentences, where each rewritten sentence clearly reflects one of the possible meanings. We categorize ambiguity into three types: lexical, syntactic, and semantic-pragmatic.

We conducted extensive experiments to investigate how LLMs handle ambiguity and found that they often exhibit fragile behavior in such scenarios. Our initial observation is that LLMs tend to confidently commit to one possible interpretation of an ambiguous sentence, which diverges from how humans typically respond to ambiguity. Furthermore, when explicitly asked to disambiguate, the models often assert with overconfidence that the sentence is ambiguous, even when it may not be. In some cases, the models demonstrate signs of \emph{overthinking} when prompted to explain ambiguous content, producing unnecessarily complex or speculative explanations.

Our analysis spans a range of open-weight LLMs, including both standard and reasoning models, from small to large scales. We performed a series of experiments involving prompt engineering and retrieval-augmented generation (RAG) across different model families and sizes. The results show that even state-of-the-art open-weight models such as DeepSeek-R1 display fragile behavior when confronted with ambiguity.
Our contributions lie in several dimensions:
\begin{itemize}
    \item The study sheds light on the semantic boundaries of LLMs, demonstrating that disambiguation remains a major challenge.
    \item This study provides a meaningful new perspective for evaluating the trustworthiness of LLMs and related systems.
    \item In the discourse of NLP research, we present and open-source a new benchmark for ambiguity detection and understanding. Meanwhile, we conduct extensive experiments and analyses to investigate how LLMs behave when faced with ambiguous sentences. Furthermore, we propose a solution to improve the robustness of LLMs in such scenarios.
\end{itemize}

This work serves as a call for the community to pay closer attention to the fragility of LLMs in the face of ambiguity, and a message of caution for industry applications concerned with the trustworthiness of LLM-based AI systems to help prevent potentially catastrophic consequences.

\section{Chinese textual ambiguity benchmark for LLMs}
\subsection{Task introduction}
Ambiguity is ubiquitous and inevitable in human language, yet large language models (LLMs) rely on natural language instructions to interface with users. Given this, understanding how LLMs behave with ambiguity is essential. In this work, we focus on two core tasks: ambiguity detection and ambiguity interpretation. In the context of NLP, the first task evaluates whether an LLM can identify if a sentence is ambiguous, formulated as a binary classification problem. The second examines whether an LLM can capture latent ambiguity and generate all plausible interpretations, framed as a conditional generation task. To support this study, we introduce a new human-annotated benchmark for ambiguity detection and interpretation in Chinese. While we acknowledge that human language lacks precision and annotations may not represent absolute ground truth, our goal is to analyze the behavior of LLMs in the face of ambiguity, highlight discrepancies with human judgments, and offer a new perspective on LLM evaluation. Through the lens of these tasks, we aim to investigate the following research questions:

\begin{itemize}
\item \textbf{RQ1}: To what extent does an LLM differ from human annotators in identifying ambiguous narratives?
\item \textbf{RQ2}: How does an LLM perform when explaining the meaning of a sentence that contains ambiguity?
\item \textbf{RQ3}: How does an LLM interpret the meaning of a sentence when it is explicitly informed that the sentence is ambiguous?
\end{itemize}

\subsection{Dataset creation}
In this work, we employ human annotators to construct ambiguous sentences along with their corresponding disambiguated versions. Annotators are also asked to provide all plausible interpretations of each ambiguous sentence. The sentence construction is grounded in real-world scenarios and everyday contexts, and the data are sourced through original writing, commonly used spoken expressions, online searches, and AI-assisted generation. The quality of the annotations is assessed by the annotators. All annotators are native Chinese speakers with qualifications sufficient for admission to graduate-level programs in science and engineering. We include only sentences that remain highly ambiguous and cannot be easily disambiguated based on human annotators' judgments.

After sentence collection and annotation, we further categorize ambiguity into three main levels: lexical, syntactic, and semantic-pragmatic, following established studies in Chinese linguistics. Within the lexical category, we further distinguish homonymy, polysemy, and part-of-speech ambiguity. Within syntactic ambiguity, we include both structural and syntax–semantics ambiguity. For semantic-pragmatic ambiguity, we identify four subtypes: speech act ambiguity, conversational implicature, deixis ambiguity, and sociocultural ambiguity. The ambiguity categories and label statistics are presented in Table \ref{tab:ambiguity-taxonomy} with examples.

\begin{CJK}{UTF8}{gbsn}

\begin{table*}[ht]
\caption{Categorization of Chinese ambiguity into lexical, syntactic, and semantic-pragmatic levels, each with multiple English interpretations based on contextual usage. Each category and sub-category is accompanied by its corresponding statistics.}
\label{tab:ambiguity-taxonomy}
\centering
\resizebox{\textwidth}{!}{
\begin{tabular}{|l|l|p{5.5cm}|p{5.5cm}|}
\hline
\textbf{Category} & \textbf{Sub-category} & \textbf{Ambiguity Example in Chinese} & \textbf{English Translation} \\
\hline
\multirow{3}{*}{\centering Lexical (218)} 
& Polysemy (152) & 吃完饭，他冷冷地说：“这顿饭先记着，回头我们再算账。” & “Let's keep this in mind and settle the bill later.” / “Let's keep this in mind — I'll get even with you later.” (“算账” can mean settling payment or seeking revenge) \\
\cline{2-4}
& Homonymy (27) & 小明走在公园里，赞叹枝头上的杜鹃很漂亮。 & Xiao Ming walked in the park and admired how beautiful the cuckoo was on the branch. / ... admired how beautiful the azalea was on the branch. (“杜鹃” can mean a bird or a flower) \\
\cline{2-4}
& Part-of-Speech (39) & 民警来到现场勘察，发现这个门没有锁。 & The police arrived and found that the door didn't have a lock. / ... found that the door hadn't been locked. (“锁” as noun vs. verb) \\
\hline
\multirow{2}{*}{\centering Syntactic (327)} 
& Structural Ambiguity (261) & 接到紧急通知后，领导简单地宣布：我们需要组织人员。 & Upon receiving the emergency notice, the leader briefly announced: “We need to organize the personnel.” / “We need the staff responsible for organization.” (“组织人员” can be verb-object or compound noun) \\
\cline{2-4}
& Syntax–Semantics (66) & 女儿在日记中写道：我恨她对我的刻薄不容忍。 & The daughter wrote in her diary: “I hate that she cannot tolerate my harshness.” / “I hate her harshness and intolerance toward me.” (ambiguity in scope of negation and attribution) \\
\hline
\multirow{4}{*}{\centering Semantic-Pragmatic (357)} 
& Speech Act (101) & 病房里护士问病人："你能把窗户关上吗？" & “Can you close the window?” (literal inquiry about capability) / “Please close the window.” (polite indirect request) \\
\cline{2-4}
& Conversational Implicature (82) & 聚餐时有人提议喝酒，小王说："你们可真懂我。" & At dinner, someone suggested drinking. Xiao Wang said, “You really understand me.” (literal agreement) / “You really don't understand me at all.” (ironic/sarcastic implication) \\
\cline{2-4}
& Deixis Ambiguity (51) & 医务室王医生突然插话说：其实开刀的是他父亲。 & Dr. Wang suddenly interjected: “Actually, it was his father who had the surgery.” / “Actually, his father was the one who performed the surgery.” (“开刀” can mean to undergo or to perform surgery) \\
\cline{2-4}
& Sociocultural Ambiguity (123) & 相亲时介绍人表示，对方孩子特别老实。 & During a blind date, the matchmaker said, “Their child is very well-behaved.” (praise) / “Their child is overly obedient and lacks personality.” (implied criticism — “老实” has dual connotations in social contexts) \\
\hline
\end{tabular}
}
\end{table*}

\end{CJK}

\section{Experiment and result}

To evaluate the performance of different models on our three designed experimental tasks, we selected eight representative large language models, covering a range of scales and architectural characteristics:\\

\noindent\textbf{Qwen3 Series Models:} This includes Qwen3-4B, Qwen3-14B, Qwen3-32B, and Qwen3-235B-A22B, corresponding to 4B, 14B, 32B, and 235B parameters, respectively. Among them, Qwen3-235B-A22B is specifically optimized for reasoning-intensive tasks and shows outstanding performance in complex reasoning scenarios \cite{yang2025qwen3}.\\

\noindent\textbf{Gemma2 Series Models:} Developed by Google, these instruction-tuned models include Gemma2-2B-it, Gemma2-9B-it, and Gemma2-27B-it, with 2B, 9B, and 27B parameters, respectively. These models excel in instruction following and dialogue tasks \cite{team2024gemma}.\\

\noindent\textbf{DeepSeek-R1 Model:} A large-scale model with 671B parameters, deeply optimized for reasoning tasks, showing strong capabilities in mathematical reasoning and logical analysis \cite{guo2025deepseek}.\\

To ensure reproducibility and comparability, we split the dataset into training, development, and test sets in a 70/15/15 ratio, using stratified sampling based on ambiguity subcategories to ensure consistent distribution of ambiguity types across subsets. All input texts were pre-processed for standardization, including removing extra spaces and unifying punctuation formats, to ensure input consistency. Model outputs were also post-processed, including formatting, answer extraction, and consistency checks. All tasks used the same data split strategy to guarantee the comparability of experimental results.

In order to investigate the models' ability to identify and understand potential ambiguity, we design two experimental conditions. The first is the non-explicit prompt condition (\textbf{Direct Interpretation}), where the prompt does not explicitly indicate that the input sentence may be ambiguous.  The second is the explicit ambiguity prompt condition (\textbf{Prompted Disambiguation}),  where the prompt explicitly states that the input sentence contains ambiguity.


\subsection{Experimental Tasks}

Based on the aforementioned research questions and task formalization, we adopted a structured experimental design and completed three experimental tasks, systematically addressing the three core issues in Chinese ambiguity processing: ambiguity detection, ambiguity understanding, and end-to-end detection and understanding. For evaluation, we used accuracy, precision, recall, and F1 score as the main metrics. Given the imbalanced distribution of ambiguous sentences in real-world corpora, we placed particular emphasis on F1 score and recall. We constructed a multi-dimensional, multi-level evaluation framework to comprehensively reflect the performance of different methods on Chinese ambiguity processing tasks.

\subsubsection{Ambiguity Detection Task}

The core goal of the ambiguity detection task is to perform binary classification on a given Chinese sentence, i.e., to determine whether the sentence contains ambiguity. In this task, the provided sentences may or may not be ambiguous, and the model needs to make its own judgment and respond with 'yes' or 'no'. This task forms the foundation of the entire ambiguity processing pipeline. The evaluation is based on standard binary classification metrics, including accuracy, precision, recall, and F1 score.

\subsubsection{Ambiguity Understanding Task}

The ambiguity understanding task is a further extension based on ambiguity detection, requiring the model to, given a Chinese sentence (with or without ambiguity), complete three sub-tasks: ambiguity source localization, multiple interpretation generation, and disambiguated sentence generation. Specifically, the model needs to identify words, phrases, or syntactic structures that may cause ambiguity and mark their positions; then, based on these sources, generate all reasonable and semantically coherent interpretations (if ambiguity exists, at least two different interpretations should be provided); finally, for each interpretation, generate a corresponding disambiguated sentence by adding context, replacing words, or adjusting structure to eliminate ambiguity. To comprehensively evaluate model capability, we designed two experimental conditions: (1) directly prompting the model to explain possible meanings without indicating whether ambiguity exists, to assess the model's overall detection and understanding ability; (2) explicitly indicating that the sentence contains ambiguity, to focus on the model's understanding and generation ability. This task places higher demands on the model's linguistic analysis, understanding, and generation capabilities. Evaluation uses exact match (EM), recall, and set F1 to assess the quality of generated interpretations, effectively reflecting model performance in multi-interpretation generation.

\subsubsection{End-to-End Detection and Understanding Task}

The end-to-end task represents the highest level of ambiguity processing. Given a raw sentence, the model must first perform ambiguity detection and ambiguity type recognition, then combine the detection results with other prompting strategies (such as chain-of-thought, RAG, etc.) to form composite prompts, guiding the large model to complete ambiguity understanding and disambiguation, and automatically output multiple interpretations and disambiguated sentences. The detection results at each stage serve as prompt conditions for subsequent reasoning, with all steps integrated into a single pipeline, achieving fully automated processing from raw input to final output without human intervention. This setup not only closely simulates real-world application scenarios, but also greatly increases task complexity. Evaluation uses joint metrics, comprehensively considering detection accuracy and understanding quality, providing a quantitative assessment of overall task performance.

We clarify that all accuracy-related metrics are used solely to measure alignment with human annotations, rather than to define any absolute ground truth. We do not claim an objective standard for determining whether a sentence is ambiguous, as natural language is inherently ambiguous and human interpretations can vary significantly. This limitation should be acknowledged when interpreting the results.

\subsection{Detection Methods}

The detection methods include both transformer-based text classifiers and large language model prompting. By observing the changes in model performance under different prompting strategies, we analyze how the design of prompts affects the model's ability to handle ambiguity.
\begin{table}[h!]
\caption{Ambiguity detection performance across different LLMs. \textbf{Bolded} scores represent the best performance, and \underline{underlined} scores indicate the second-best results. \textsuperscript{\dag} These models are optimized for reasoning tasks and have reasoning explicitly enabled.}
\label{tab:ambiguity-detection}
\centering
\resizebox{\columnwidth}{!}{
\begin{tabular}{llcccc}
\hline
\textbf{Model} & \textbf{Params} & \textbf{Accuracy} & \textbf{Precision} & \textbf{Recall} & \textbf{Macro-F1} \\
\hline
BERT-ft       & 109M         & \textbf{94.70} & \textbf{94.16} & \textbf{89.58} & \textbf{91.81} \\
\hline
\multirow{3}{*}{Gemma2}
           & 2B         & 46.06     & 49.60     & 49.57     & 45.77 \\
           & 9B         & 38.19     & 52.24     & 51.18     & 35.85 \\
           & 27B        & 43.20     & 50.53     & 50.50     & 43.14 \\
\hline
\multirow{3}{*}{Qwen3} 
           & 4B         & 63.96     & 53.32    & 51.93     & 50.24 \\
           & 14B        & 58.95     & 54.63     & 54.91     & 54.65 \\
           & 32B        & 55.85     & 56.66     & 57.58     & 54.79 \\
\hline
Qwen3\textsuperscript{\dag}     & 235B-A22B  & 43.68     & 54.97     & 53.91     & 43.08 \\
DeepSeek-R1\textsuperscript{\dag}   & 671B-A37B   & \underline{65.63}     & \underline{62.41}     & \underline{63.48}    & \underline{62.62} \\
\hline
\end{tabular}
}

\end{table}

\subsubsection{Pretrained Transformer-based Text Classifier}
Transformer-based pre-trained language models, such as BERT \cite{devlin2019bert}, RoBERTa \cite{liu2019roberta}, and XLNet \cite{yang2019xlnet}, have demonstrated strong performance across a wide range of text classification tasks. Given their effectiveness in sentence-level modeling \cite{tian2020skep, zhang2021none, zhang-etal-2024-positive} and passage-level discrimination \cite{chang2020taming, huang2025hybrid} in applied classification scenarios, we adopt Transformer-based classifiers as our foundation.
As a baseline, we used the pre-trained language model \texttt{hfl/chinese\allowbreak-roberta\allowbreak-wwm\allowbreak-ext} \cite{cui-etal-2020-revisiting} as the classifier backbone. This model is based on the RoBERTa architecture and has been specifically optimized for Chinese, achieving strong performance in various NLP tasks. We further fine-tune the model with a binary classification objective to distinguish between ambiguous and unambiguous sentences. 

We added a classification head to the model and fine-tuned it for binary classification. Regarding feature engineering, in addition to textual input, we incorporated linguistic features such as sentence length, POS tag sequences, and syntactic tree depth to enhance the model's sensitivity to Chinese ambiguity. These features were fused with the main model output via additional embedding layers.

\begin{table*}[h!]
\caption{Macro-F1 performance on ambiguity detection using different prompting strategies. \textbf{Bolded} scores represent the best performing model under each method, while \underline{underlined} scores indicate the best performing method for each model. \textsuperscript{\dag} Reasoning-enabled models.}
\label{tab:ambiguity-prompt-macro-f1}
\centering
\resizebox{0.8\textwidth}{!}{
\begin{tabular}{ll|ccccccc}
\hline
\textbf{Model} & \textbf{Params} 
& \textbf{Direct Prompt} 
& \textbf{Few-shot} 
& \textbf{Knowledge} 
& \textbf{CoT} 
& \textbf{CoT + FS} 
& \textbf{RAG + FS} \\
\hline
\multirow{3}{*}{Gemma2}
 & 2B         & 45.77 & 39.58 & 40.40 & 34.50 & \underline{51.95} & 46.69 \\
 & 9B         & 35.85 & 32.09 & 38.75 & 37.32 & 41.74 & \underline{52.95} \\
 & 27B        & 43.14 & 40.75 & 42.65 & 36.32 & 44.61 & \underline{56.12} \\
\hline
\multirow{3}{*}{Qwen3} 
 & 4B         & 50.24 & 43.02 & 50.95 & 46.86 & 47.71 & \underline{58.05} \\
 & 14B        & 54.65 & 53.81 & 52.11 & 42.24 & 52.51 & \underline{60.83} \\
 & 32B        & 54.79 & 55.23 & 55.00 & 44.20 & 55.72 & \underline{69.57} \\
\hline
Qwen3\textsuperscript{\dag}     & 235B-A22B  &43.08  & 55.46 & 57.68 & 53.25 & 59.35 & \underline{74.41} \\
DeepSeek-R1\textsuperscript{\dag}   & 671B-A37B   &\textbf{62.62}  & \textbf{63.94} & \textbf{62.63} & \textbf{55.20} & \textbf{65.16} & \textbf{\underline{87.01}} \\
\hline
\end{tabular}
}

\end{table*}

\begin{table*}[h!]
\caption{Performance on ambiguity meaning understanding task. Models are evaluated in two settings: \textbf{Direct Interpretation} (without disambiguation prompt) and \textbf{Prompted Disambiguation} (with explicit disambiguation prompt). Metrics include Set F1, Recall, and Exact Match (EM). \textbf{\(\Delta\) Set F1 / \(\Delta\) Recall} shows the improvement from prompting. \textsuperscript{\dag} Reasoning-enabled models.}
\label{tab:ambiguity-meaning-comparison}
\centering
\resizebox{0.8\textwidth}{!}{
\begin{tabular}{ll|ccc|ccc|c}
\hline
\textbf{Model} & \textbf{Params} 
& \multicolumn{3}{c|}{\textbf{Direct Interpretation}} 
& \multicolumn{3}{c|}{\textbf{Prompted Disambiguation}} 
& \textbf{Difference} \\
\cline{3-8}
& & \textbf{EM} & \textbf{Recall} & \textbf{Set F1} 
  & \textbf{EM} & \textbf{Recall} & \textbf{Set F1} 
  & \textbf{\(\Delta\) Set F1 / \(\Delta\) Recall} \\
\hline
\multirow{3}{*}{Gemma2}
 & 2B  & 0.00 & 26.65 & 40.49 & 0.00 & 27.21 & 41.18 & 0.69 / 0.55 \\
 & 9B  & 0.00 & 30.33 & 44.71 & 0.00 & 29.78 & 43.92 & -0.78 / -0.55 \\
 & 27B & 0.00 & 31.62  & 46.37 & 0.00 & 31.07 & 45.69 &  -0.69 / -0.55 \\
\hline
\multirow{3}{*}{Qwen3} 
 & 4B  & 0.00 & 31.99 & 46.86 & 0.00 & 32.17  & 47.16  & 0.29 / 0.18 \\
 & 14B & 0.00 & 33.64 & 48.87 & 0.00 & 31.62 & 46.27 & -2.59 / -2.02 \\
 & 32B & 0.00 & 32.17 & 47.16 & 0.00 & 31.07 & 45.88 & -1.27 / -1.10 \\
\hline
Qwen3\textsuperscript{\dag} & 235B-A22B  & 0.00 & 36.40 & 51.67 & 0.00 & 37.32 & 52.65 & 0.98 / 0.92 \\
DeepSeek-R1\textsuperscript{\dag} & 671B-A37B & 0.00 & \textbf{39.71} & \textbf{55.49} & 0.00 & \textbf{40.26} & \textbf{56.18} & 0.69 / 0.55 \\
\hline
\end{tabular}
}
\end{table*}

To systematically evaluate the performance of LLMs on ambiguity detection, we designed a series of experiments to investigate the impact of model scale and prompting strategy on task performance.

\subsubsection{Large Language Model Prompt Learning}

Given the strong performance of large language models in complex reasoning tasks, we designed six different prompting strategies to systematically evaluate their effectiveness in ambiguity detection:

\noindent\textbf{(1) Direct Prompting:} In the most basic method, the model receives the input sentence and directly answers \emph{yes} or \emph{no} to indicate whether there is ambiguity. The prompt template is concise and avoids introducing bias. For example: ``Please determine whether the following sentence contains ambiguity. Just answer `yes' or `no': [sentence]"

\noindent\textbf{(2) Few-shot Prompting:} To leverage in-context learning, we include three carefully selected examples in the prompt that cover both ambiguous and unambiguous sentences. These examples represent different types of ambiguity, helping the model understand the task requirements. Selection follows the principles of representativeness and diversity.

\noindent\textbf{(3) Knowledge-enhanced Prompting:} We incorporate linguistic background knowledge about Chinese ambiguity into the prompt, including definitions and characteristics of lexical, syntactic, and semantic ambiguity. This approach aims to enhance the model's theoretical understanding and improve detection accuracy and consistency.

\noindent\textbf{(4) Chain-of-Thought Prompting:} Inspired by chain-of-thought reasoning, we require the model to perform step-by-step analysis before making a final judgment. The model first analyzes the sentence structure, then identifies possible ambiguity points, and finally provides reasoning and a conclusion, improving interpretability.

\noindent\textbf{(5) Chain-of-Thought and Few-shot Combined Prompting:} This method combines the advantages of chain-of-thought reasoning and few-shot learning, providing examples with detailed analytical processes and requiring the model to follow similar reasoning patterns for new sentences.

\noindent\textbf{(6) RAG and Few-shot Combined Prompting:} {Our approach employs a RAG and few-shot prompting strategy that pre-retrieves relevant examples to construct prompt templates for guiding model reasoning. This strategy aims to address two key issues in model understanding through the guidance of semantically similar examples, thereby improving model comprehension quality: first, the tendency to select a single possible interpretation rather than all reasonable ones; second, the problem of over-interpretation and false reasoning when sufficient context is lacking.\\

\begin{table*}[ht]
\caption{Set F1 performance on ambiguity meaning understanding under different prompting strategies. Each model is evaluated in two settings: \textbf{Direct Interpretation} (no disambiguation prompt) and \textbf{Prompted Disambiguation} (with disambiguation prompt). Methods include Direct prompt, Few-shot(FS), Knowledge+Prompt, Chain-of-Thought (CoT), CoT + Few-shot, and RAG-based Few-shot. \textbf{Bolded} scores represent the best performing model under each method, while \underline{underlined} scores indicate the best performing method for each model. \textsuperscript{\dag} Reasoning-enabled models.}
\label{tab:setf1-prompt-methods}
\centering
\resizebox{\textwidth}{!}{
\begin{tabular}{ll|cccccc|cccccc}
\hline
\textbf{Model} & \textbf{Params} 
& \multicolumn{6}{c|}{\textbf{Direct Interpretation (Set F1)}} 
& \multicolumn{6}{c}{\textbf{Prompted Disambiguation (Set F1)}} \\
\cline{3-14}
& & Direct & Few-shot & Knowledge & CoT & CoT+FS & RAG-FS 
  & Direct & Few-shot & Knowledge & CoT & CoT+FS & RAG-FS \\
\hline
\multirow{3}{*}{Gemma2}
 & 2B  & 40.49 & 44.79 & 46.01 & 44.23 & 45.46 & \underline{54.67}  & 41.18 & 47.83 & 49.31 & 50.14 & 48.04 & \underline{55.15} \\
 & 9B  & 44.71 & 48.99 & 52.07 & 49.93 & 48.68 & \underline{58.46}  & 43.92 & 50.76 & 52.58 & 52.78 &  53.57& \underline{62.50} \\
 & 27B & 46.37  &50.45 & 54.74 & 51.07 & 50.26 & \underline{61.47}  & 45.69 & 53.96 & 55.03 & 53.57 & 52.26 & \underline{63.97} \\
\hline
\multirow{3}{*}{Qwen3} 
 & 4B  & 46.86 & 51.37 & 53.37 & 48.46 & 48.04 & \underline{56.62}  & 47.16 & 55.13 & 53.76 & 53.17 & 52.77 & \underline{59.93} \\
 & 14B & 48.87 & 52.97 & 54.06 & 51.21 & 53.26 & \underline{61.03}  & 46.27 & 55.70 & 56.84 & 54.35 & 56.46 & \underline{63.60} \\
 & 32B & 47.16 & 55.13 & 54.74 & 51.86 & 53.96 & \textbf{\underline{65.07}}  & 45.88 & 55.51 & 56.53 & 56.17 & 56.54 & \textbf{\underline{67.65}} \\
\hline
Qwen3\textsuperscript{\dag} & 235B-A22B  & 51.67 & 58.33 & 59.61 & 54.74 & 59.25 & \underline{63.70}  & 52.65 & 57.96 & \textbf{61.40} & 59.43 & 59.12 & \underline{61.76} \\
DeepSeek-R1\textsuperscript{\dag} & 671B-A37B & \textbf{55.49} & \textbf{61.40} & \textbf{61.31} & \textbf{57.22} & \textbf{\underline{61.75}} & 59.85  & \textbf{56.18} & \textbf{61.05} & 60.69 & \textbf{\underline{62.10}} & \textbf{59.97} & 59.63 \\
\hline
\end{tabular}
}

\end{table*}

Table \ref{tab:ambiguity-detection} demonstrates that a fine-tuned BERT model can reliably distinguish ambiguous sentences from unambiguous ones with high accuracy. These results establish a strong baseline for ambiguity detection and indicate that incorporating a lightweight classifier may be a practical and effective enhancement in meaning-sensitive applications, particularly in settings where computational efficiency is a priority. In contrast, despite their strong reasoning capabilities, the reasoning LLMs exhibit poor performance in ambiguity detection, frequently misclassifying clear, unambiguous sentences (as determined by human annotators) as ambiguous. This tendency to over-predict ambiguity weakens their practical reliability in meaning-sensitive tasks.

Through experiments, as shown in Table \ref{tab:ambiguity-prompt-macro-f1}, we observed that the effectiveness of prompting strategies in Chinese ambiguity detection heavily relies on models' intrinsic reasoning capabilities and parameter scale. For small-to-medium models, the Chain-of-Thought (CoT) method demonstrated limited effectiveness. However, performance improved significantly when supplemented with few-shot examples. This suggests that small-to-medium models lack the necessary reasoning capacity to leverage CoT strategies effectively; instead, they better grasp task fundamentals through concrete examples, thereby substantially enhancing their ability to detect ambiguity. We also found that specialized reasoning models (e.g., Qwen3 and DeepSeek-R1) excelled across all prompting strategies and could more effectively unlock their potential using the CoT+FS strategy to achieve peak performance. In contrast, non-reasoning-specialized models relied more heavily on external knowledge frameworks provided by Few-shot and Knowledge strategies, with their internal reasoning processes offering limited guidance. 

\begin{figure}
    \centering
    \includegraphics[width=1\linewidth]{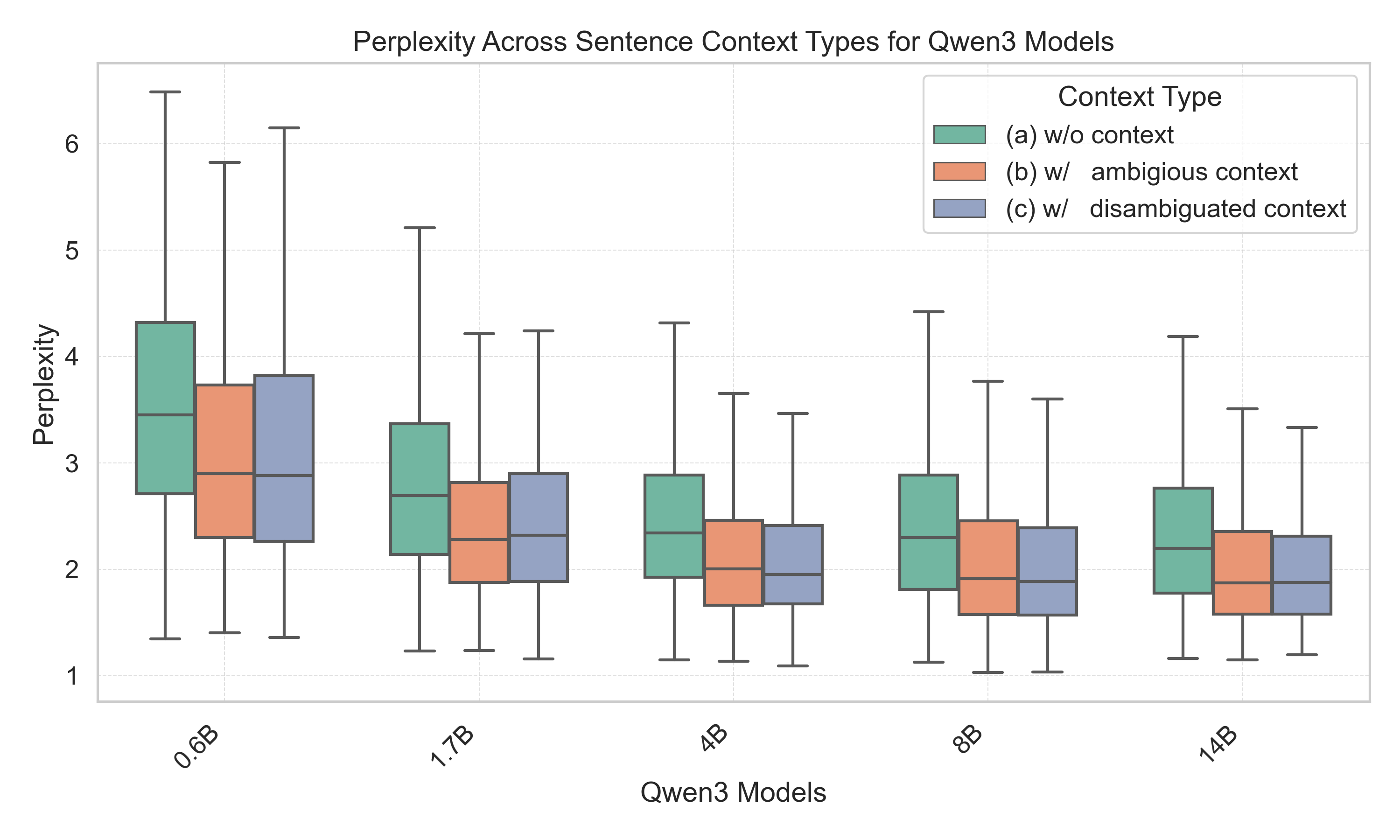}
    \caption{Perplexity scores of Qwen3 models for ambiguous sentences across three context types. When similar context is given for a sentence, no matter ambiguous or disambiguated, there is no observed significant difference in perplexity.}
    \label{fig:perplexity_comparision}
\end{figure}

Table \ref{tab:ambiguity-meaning-comparison} presents a comprehensive evaluation comparing two strategies: Direct Interpretation (asking for the meaning directly) and Prompted Disambiguation (asking for the meaning with an explicit cue that the sentence is ambiguous). The evaluation is conducted by comparing the predicted set of meanings with the gold-standard set, using exact match, recall, and set-level F1 score as metrics. The results indicate that models perform poorly on this task, and the inclusion of an ambiguity prompt does not yield consistent or reliable improvements. Given the prohibitive cost of human evaluation at scale, especially for tens of thousands of meaning-level sentence comparisons, we employ strong reasoning models to approximate this process by comparing the predicted and reference meaning sets and outputting the number of overlapping meanings, which is then used to compute the evaluation metrics.

The performance gap between the Direct Interpretation and the Prompted Disambiguation frameworks in Table \ref{tab:setf1-prompt-methods} reveals the significant impact of instruction framing on model comprehension. Under the Prompted Disambiguation framework, models consistently outperformed those using Direct Interpretation across all prompting strategies, demonstrating that explicit ambiguity-specific prompting enhances models' sensitivity to multi-interpretation scenarios. These findings provide a theoretical basis for optimizing large language models in Chinese ambiguity understanding tasks and reveal differential sensitivity to prompting strategies across models of varying scales.

\begin{figure}
    \centering
    \includegraphics[width=1\linewidth]{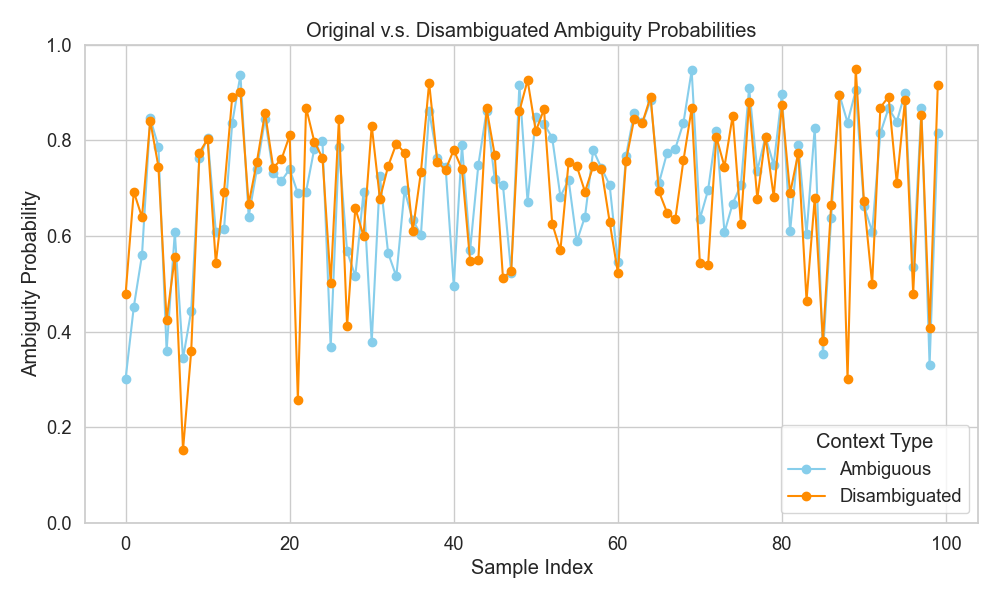}
    \caption{Comparison of ambiguity probabilities (the probability that Qwen3-8B model answers \emph{YES} relative to \emph{NO} to the question \emph{Is the sentence ambiguous or not?}) between ambiguous sentences and their disambiguated versions. The disambiguation does not consistently reduce the model's perceived ambiguity.}
    \label{fig:probs_comparision}
\end{figure}

\begin{figure*}
    \centering
    \includegraphics[width=1\linewidth]{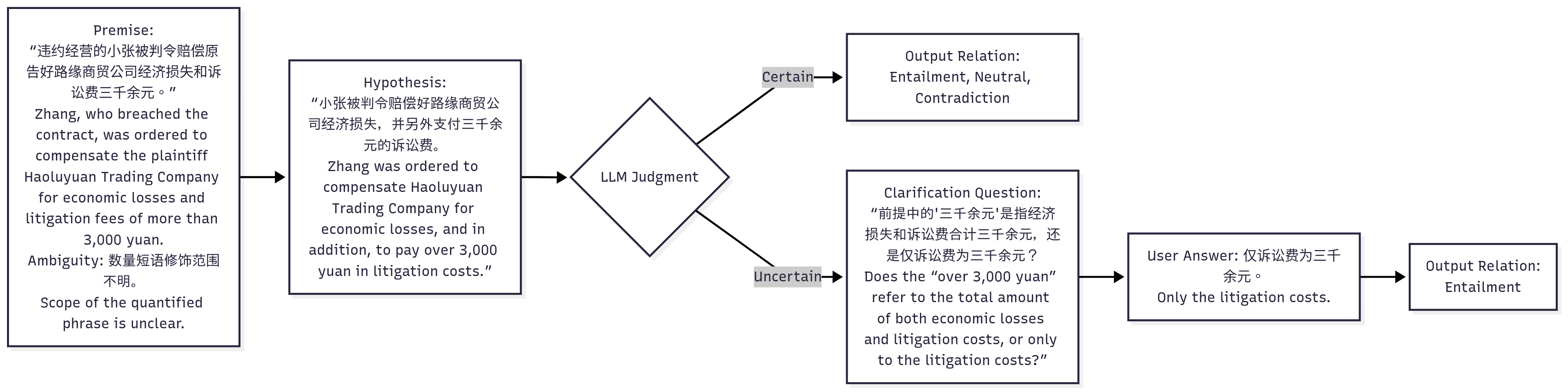}
    \caption{Example workflow for resolving ambiguity through clarification questions.}
    \label{fig:case_study_1}
\end{figure*}

\begin{figure*}
    \centering
    \includegraphics[width=0.9\linewidth]{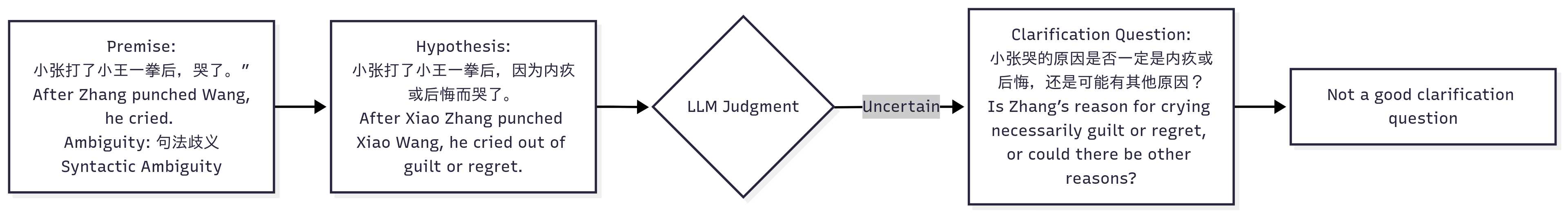}
    \caption{Illustration of an LLM's failure to generate an effective clarification question. The model incorrectly focuses on emotional reasoning (guilt or regret) rather than resolving the core syntactic ambiguity.}
    \label{fig:case_study_2}
\end{figure*}

Through evaluations on three tasks: ambiguity detection, ambiguity understanding, and end-to-end assessment, as shown in Table \ref{tab:ambiguity-prompt-macro-f1} and Table \ref{tab:setf1-prompt-methods}, we observe that model performance improves with increased parameter size in both ambiguity detection and meaning understanding. Reasoning models often perform better across different prompting methods. We also find that the RAG method enhances sensitivity to Chinese ambiguity, especially for medium-sized non-reasoning models, by helping them identify multiple interpretations using relevant examples. Moreover, larger models benefit more from RAG, suggesting that reasoning ability plays a key role in handling ambiguity. Specifically, in the ambiguity detection task, due to its relative simplicity, the RAG strategy shows significant improvements across all models; in ambiguity understanding and end-to-end evaluation, due to the increased task complexity, the improvement effects of the RAG strategy have upper limits, primarily constrained by the models' inherent reasoning capabilities. As shown in Table \ref{tab:setf1-prompt-methods}, RAG provides modest improvements for non-reasoning models, while showing limited enhancement for models with strong reasoning capabilities (such as DeepSeek-R1). This occurs because strong reasoning models rely more on internal logic rather than external prompts, and are more sensitive to retrieval noise. Although high-quality retrieval still has positive effects, low-quality retrieval may have negative impacts, exhibiting diminishing marginal returns. For medium-scale models, RAG provides additional reasoning pathways that compensate for their insufficient reasoning capabilities, while these models have sufficient capacity to process rich examples, making them the optimal range for RAG strategy application. Small models have limited capacity and difficulty fully utilizing complex examples, potentially being overwhelmed by excessive information.}



\section{Analysis and Discussion}
\subsection{Perplexity Analysis}
A language model's perplexity (PPL) on a sequence of tokens is calculated by averaging the log probability values of its predictions for each token in the sequence. Perplexity is a statistical metric that assesses a language model's ability to predict a text sequence, reflecting the model's uncertainty in assigning probabilities to upcoming tokens. While PPL is more considered as a measure that evaluates how well LLMs model text patterns, we also assume that it measures LLMs' ability to understand text.

Since PPL scores are strongly affected by the model's training data, they cannot be directly compared between different models or across different datasets. Nevertheless, if all models share the same training data, the PPL scores become more comparable. In this case, differences in perplexity can more reliably reflect variations in text understanding. Inspired by this, researchers developed log-probability-based methods to classify potentially deceptive articles \cite{zhang2020birds, lee2021towards} and check AI-generated content \cite{mitchell2023detectgpt, xu2024detecting}.

In this study, we compare the PPL scores of a set of Qwen3 models on our benchmark to evaluate their relative certainty and predictive performance. Since all these models share the same training data and vocabulary, the perplexity values are directly comparable to some degree. This comparison can provide insights into how confidently each model handles the benchmark's input sequences.

For each sample in the benchmark, we measure a triplet of PPL scores: (a) the PPL of the ambiguous sentence without preceding or following context; (b) the PPL of the ambiguous sentence with ambiguous context; (c) the PPL of the ambiguous sentence with disambiguated context. We filter out samples whose (b) and (c) versions differ substantially in length, ensuring that the PPL scores are more comparable.

The results are shown in Figure \ref{fig:perplexity_comparision}. We observe that sentences with context generally have lower perplexity than those without context. However, when the provided context is similar in both length and semantic meaning, regardless of whether they are ambiguous or disambiguated, there is no significant difference in perplexity between them. This observation suggests that PPL scores may not serve as a reliable signal for LLMs' ambiguity understanding ability. We also note that larger models tend to have lower perplexity scores, suggesting that they are more confident in their understanding of those ambiguous sentences.

As part of analyzing the decoding dynamics of Qwen3 models under conditioned inputs, we evaluate their token-level log-probability assignments on pairs of ambiguous and disambiguated sentences. For each sentence, we explicitly ask whether it is ambiguous to assess the model's inherent sensitivity to ambiguity.
Based on prior assumptions, we hypothesized that ambiguous sentences would elicit higher probabilities for a \emph{YES} response compared to their disambiguated counterparts. However, as shown in Figure \ref{fig:probs_comparision}, no clear or consistent pattern was observed. This result suggests that log-probabilities may not serve as a reliable signal for detecting ambiguity, cross-validating our earlier observation that large language models exhibit limited awareness of linguistic ambiguity in Chinese text.

\subsection{Probing Ambiguity via Clarification Questioning}
To further investigate the model's robustness against Chinese textual ambiguity, we propose an evaluation method inspired by Natural Language Inference (NLI) framing.

Every premise contains an ambiguous expression with two possible interpretations (A and B). Then, three hypotheses are generated: \textbf{Entailment}: is inferable from the premise with interpretation A. \textbf{Neutral}: Remains ambiguous, committing to neither A nor B. \textbf{Contradiction}: Supports Interpretation B and logically contradicts A \cite{liu2023we}.

Figure \ref{fig:case_study_1} illustrates the step-by-step process by which an LLM addresses semantic ambiguity in an NLI scenario. When given an ambiguous premise, the model may fail to make a definitive inference judgment. It generates a clarification question to explicitly resolve the ambiguity. Once the user provides a disambiguating answer, the model determines the inference relation: entailment, contradiction, or neutral. This process provides a way to evaluate whether the model has correctly identified and understood the ambiguity. By leveraging joint reasoning to identify the minimal conditions needed for clarification or decision-making, our analysis method is also conceptually similar with existing work on explanation through factual and counterfactual analysis \cite{chen2025joint}.



Figure \ref{fig:case_study_2} illustrates a case where the LLM correctly detects ambiguity, but misidentifies the source of ambiguity. Instead of focusing on the syntactic uncertainty (i.e., who cried), the model assumes the ambiguity regarding the reason for crying. As a result, it generates a clarification question that is misaligned with human intuition and fails to resolve the key ambiguity.

\section{Related Work}
Disambiguation has been an extensively studied research topic in NLP, as ambiguity is inherently present in human language and communication. Traditional machine learning-based NLP approaches have primarily focused on word sense disambiguation \cite{navigli2009word, bevilacqua2021recent}, employing knowledge-based methods, vector-based 1-nn classifiers, token taggers, and sequence taggers to resolve lexical ambiguity.
Ambiguity detection has also been thoroughly explored in the literature. \cite{gleich2010ambiguity} developed a taxonomy for classifying ambiguity and created POS-based and rule-based tools to detect ambiguity in requirement documents. \cite{ferrari2019nlp} trained word embeddings on domain-specific corpora and compared cross-domain term representations to automatically identify semantic ambiguities.
Large Language Models (LLMs) demonstrate exceptional capabilities in natural language understanding and reasoning tasks. Compared to early transformer models, LLMs exhibit superior performance and flexibility in comprehending and solving multiple-choice questions across diverse subjects including history, science, and mathematics, as demonstrated on benchmarks such as MMLU \cite{hendrycks2020measuring}, MMLU-Pro \cite{wang2024mmlu}, GPQA \cite{rein2024gpqa}, and AIME \cite{balunovic2025matharena}. LLMs also excel across multiple dimensions of language understanding, including commonsense reasoning \cite{krause2023commonsense} and interpretation of abstract concepts \cite{zhang2024interpreting}, and they even extend beyond the natural language domain, supporting tasks such as coding \cite{zhang2023unifying}, recommendation \cite{wu2024survey}, forecasting, and anomaly detection \cite{su2024large}. However, existing reviews \cite{huang2025unmasking, yi2025challenges} indicate that although LLMs demonstrate strong performance in language understanding tasks, they remain limited in their ability to capture fine-grained semantic nuances.
Nevertheless, ambiguity remains a fundamental linguistic phenomenon that cannot be entirely overcome and has garnered significant attention from the research community. \cite{liu2023we} presents an early work identifying limitations of LLMs in ambiguity understanding, and proposes AMBIENT, an English benchmark of ambiguous sentences. \cite{shi-etal-2025-ambiguity} specifically investigated ambiguity handling in questions to enhance LLM performance when confronted with ambiguous inputs. \cite{keluskar2024llms} explored improvements in LLM ambiguity handling for open-world question answering through simple prompt rewriting and context augmentation. \cite{mehrparvar2024detecting} examined ambiguity detection mechanisms in LLMs. CLAMBER \cite{zhang-etal-2024-clamber} addresses ambiguity challenges in query intention understanding and information clarification requirements for LLMs in retrieval tasks. Although these studies focus on enhancing LLM performance in specific applications, they do not examine the fundamental language understanding behaviors of LLMs when processing ambiguous content. In this work, we use Chinese as a case study to investigate how LLMs encounter and handle ambiguity with specific scenes, thereby providing meaningful insights for future research on ambiguity processing in LLMs.

\section{Conclusion}
In this work, we examine the fragility of large language models (LLMs) when handling textual ambiguity through Chinese oral input. We created a benchmark consisting of 900 ambiguous sentences with context across 9 categories, paired with corresponding disambiguation sentences. Our findings reveal that state-of-the-art open-weight LLMs still struggle with ambiguity detection and understanding.
Specifically, we observe three key issues. First, LLMs exhibit overconfidence when classifying sentences as ambiguous in detection tasks. Second, LLMs fail to effectively identify possible alternative meanings from ambiguous statements. Third, when explicitly prompted to understand ambiguous meanings, LLMs tend to overthink and generate meanings that are far-fetched compared to human interpretation.
Our comprehensive experiments and analyses demonstrate several important findings. Models with more parameters perform better on these tasks, and reasoning-enhanced models show improved performance in both detection and understanding. Most notably, adding examples through retrieval-augmented generation (RAG)  proves to be the most effective approach for improving both detection and understanding tasks.
We also analyzed model behavior by examining perplexity differences between ambiguous and disambiguated sentences. Additionally, we explored ambiguity through probing techniques using clarification questions with case studies.
This work provides a novel perspective on LLM trustworthiness and serves as a call for the community to address this inherent issue in LLMs and exercise caution in practical applications.
For future work, we plan to conduct fine-grained analysis within different categories of ambiguity and develop lightweight, effective methods to mitigate these problems.

\begin{acks}
We thank the anonymous reviewers for their valuable comments and suggestions.
\end{acks}

\bibliographystyle{ACM-Reference-Format}
\bibliography{sample-base}

\appendix
\section{Implementation details}
\subsection{Choice of Ambiguity Detection Model}
To better handle this Chinese-specific task, we consider using language models that are pretrained on Chinese corpora and tasks \cite{zhang2021cpm, sun2019ernie, cui-etal-2020-revisiting}. Among them, we select \texttt{hfl/chinese\allowbreak-roberta\allowbreak-wwm\allowbreak-ext} \cite{cui-etal-2020-revisiting}, a RoBERTa-based model specifically designed for Chinese.
Unlike standard BERT models that apply subword-level masking, this model adopts whole-word masking (WWM), meaning that it masks entire Chinese words during pretraining. Since Chinese words often consist of multiple characters, WWM enables the model to learn more meaningful word-level representations. This property is particularly beneficial for identifying sentence-level ambiguity, where subtle differences in phrasing can lead to different interpretations.

\subsection{Detection Model Training Procedure}

To fine-tune the model for our task, we use the training set of the manually annotated samples from the dataset. Each ambiguous sentence is paired with two disambiguated versions that preserve the original meaning while removing the ambiguity. Ambiguous and disambiguated sentences are labeled as 1 and 0, respectively. This structure provides a semantic contrast between positive and negative examples.

To enhance the input representation, we added linguistic features to each sentence. Specifically, we appended a word-segmented version of the sentence using \texttt{jieba} and the corresponding part-of-speech (POS) tags. The final input format contains both lexical and syntactic cues, aiding the model in better understanding structural aspects of Chinese that are often associated with ambiguity.

For training, the model configuration included a learning rate of 2e-5, a batch size of 16, 5 training epochs, Adam optimizer, and a linear learning rate decay schedule. Early stopping was used to prevent overfitting, halting training if validation performance did not improve for 3 consecutive epochs. We set a random seed for reproducibility and used CUDA to accelerate the computation. We applied gradient clipping (with \texttt{max\_norm=1.0}) to avoid exploding gradients, and used a linear learning rate scheduler with warm-up steps. The early stopping strategy was implemented based on the validation F1 score, with a patience of three epochs.

To reduce performance variance and improve robustness, stratified K-fold cross-validation was used during training. Additionally, we applied automated hyperparameter tuning using the Optuna framework. The search space included batch size, learning rate, and weight decay, and each configuration was evaluated based on the cross-validation F1 score. This approach allowed us to identify a better combination of parameters with minimal manual tuning.

\subsection{Choice of LLMs}
In this work, we focus exclusively on open-weight LLMs for our experiments. While proprietary models have demonstrated strong language understanding capabilities, their APIs and chat interfaces function as black boxes, making it unclear whether additional components beyond the model weights influence the outputs. This lack of transparency may affect experimental validity and reduce reproducibility. Therefore, we selected open-weight models Gemma 2, Qwen 3, and DeepSeek R1 for our study, with Qwen and DeepSeek R1 in particular being developed by Chinese researchers and showing strong performance on tasks in Chinese. Using open-weight models for benchmarking is also a reasonable practice in the research community \cite{mo-etal-2024-trustworthy, qu2025prompt}.

\section{Author Contributions}
Xinwei Wu contributed to the creation of the benchmark dataset and conducted experiments on ambiguity detection and ambiguity understanding across multiple large language models.

Hongyu Liu contributed to a portion of the benchmark dataset development and conducted comparative experiments with result evaluation for Retrieval-Augmented Generation (RAG) methodology in Chinese ambiguity processing. 

Haojie Li contributed to the creation of the benchmark dataset and implemented the full pipeline for data processing and disambiguation, including data cleaning, feature engineering, model training and optimization, and experimental evaluation.

Xinwei Wu, Hongyu Liu, and Haojie Li contributed equally to this work and share first authorship.

Xinyu Ji contributed to the creation of the benchmark dataset and conducted evaluations of ambiguity detection and ambiguity understanding across multiple large language models. All contributions were made in a personal capacity and do not reflect the views of her employer.

Ruohan Li contributed to the creation of the benchmark dataset and conducted an evaluation via clarification questioning.

Yule Chen conducted the perplexity analysis and contributed to the creation of the benchmark dataset.

Yigeng Zhang provided overall supervision and mentored the team across all stages of the project. Yigeng Zhang participated in this project as a volunteer mentor supporting early-stage researchers. All contributions were made in his personal capacity and do not represent the views of his employer.

\end{document}